\documentclass{article}
\usepackage{ijcai24}

\usepackage{times}
\usepackage{soul}
\usepackage{url}
\usepackage[hidelinks]{hyperref}
\usepackage[utf8]{inputenc}
\usepackage[small]{caption}
\usepackage{graphicx}
\usepackage{amsmath}
\usepackage{amsthm}
\usepackage{booktabs}
\usepackage{algorithm}
\usepackage{algorithmic}
\usepackage[switch]{lineno}

\usepackage{multirow}
\usepackage{tabularx}

\title{CIC: A Framework for Culturally-Aware Image Captioning}

\author{
    Youngsik Yun \textsuperscript{\rm1} and Jihie Kim \textsuperscript{\rm2} \thanks{indicates corresponding authors.}
    \affiliations
    \textsuperscript{\rm 1} Department of Computer Science and Artificial Intelligence, Dongguk University\\
    \textsuperscript{\rm 2} Division of AI Software Convergence, Dongguk University
    \emails
    yys3606@dgu.ac.kr, jihie.kim@dgu.edu
}

\begin{document}

\maketitle

\begin{abstract}
Image Captioning generates descriptive sentences from images using Vision-Language Pre-trained models (VLPs) such as BLIP, which has improved greatly. However, current methods lack the generation of detailed descriptive captions for the cultural elements depicted in the images, such as the traditional clothing worn by people from Asian cultural groups. In this paper, we propose a new framework, \textbf{Culturally-aware Image Captioning (CIC)}, that generates captions and describes cultural elements extracted from cultural visual elements in images representing cultures. Inspired by methods combining visual modality and Large Language Models (LLMs) through appropriate prompts, our framework (1) generates questions based on cultural categories from images,  (2) extracts cultural visual elements from Visual Question Answering (VQA) using generated questions, and (3) generates culturally-aware captions using LLMs with the prompts. Our human evaluation conducted on 45 participants from 4 different cultural groups with a high understanding of the corresponding culture shows that our proposed framework generates more culturally descriptive captions when compared to the image captioning baseline based on VLPs. Resources can be found at \url{https://shane3606.github.io/cic}.
\end{abstract}

\section{Introduction}
AI applications are for everyone. With the growing utilization of AI systems and applications in the real world, it has become increasingly important to develop applications that ensure fairness. It is important to ensure that the AI applications do not present discriminatory behaviors toward specific groups or populations~\cite{mehrabi2021survey}. Culture distinguishes certain groups and represents the way of life of people in the collective group~\cite{hershcovich2022challenges}. Therefore, As AI applications advance, taking appropriate consideration of the cultures of various groups is crucial~\cite{yin2023givl,li2023cultural}.  

\begin{figure}
    \centering
    \includegraphics[width=1\linewidth]{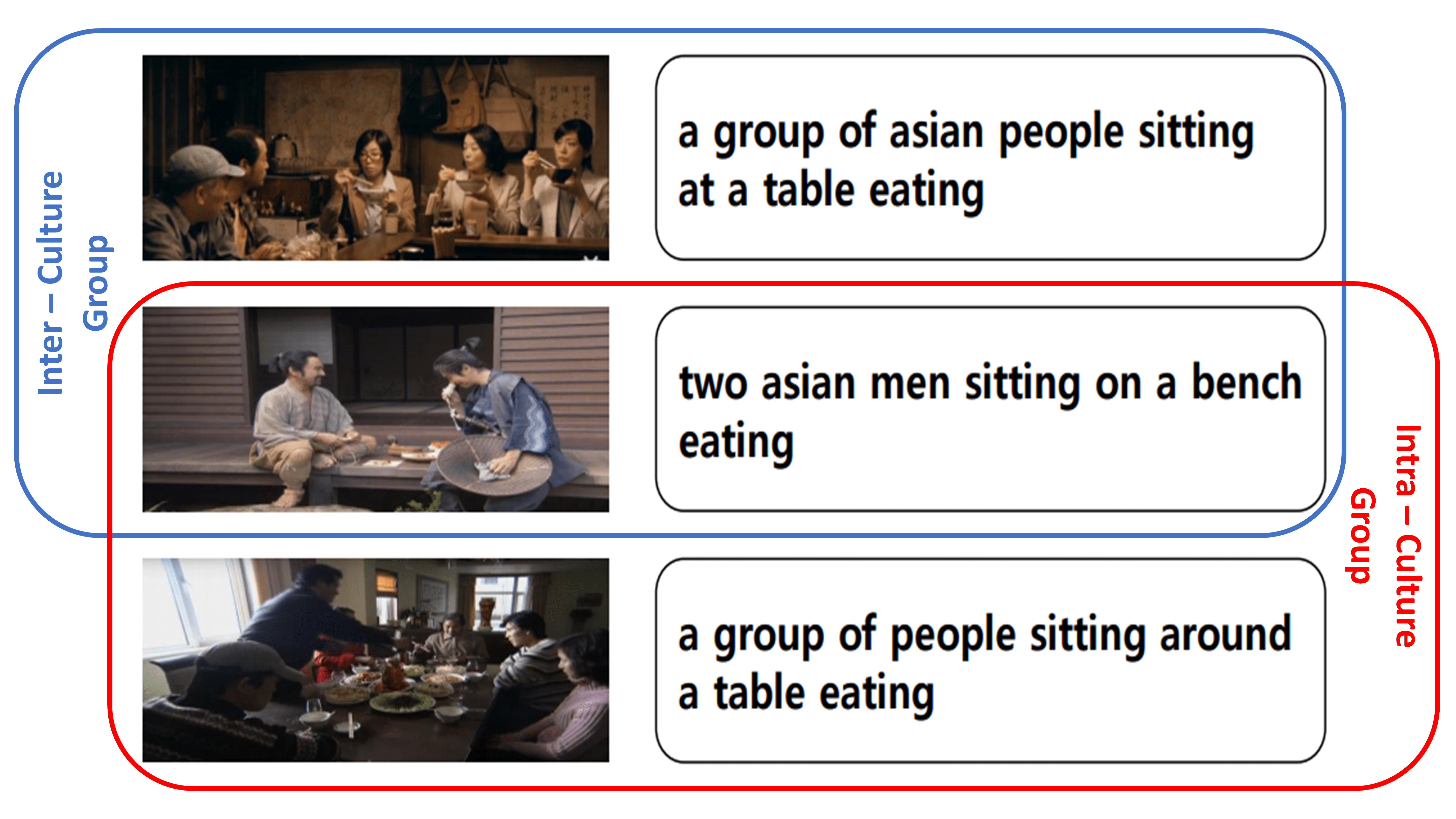}
    \caption{Comparison of generated captions from VLPs between the same culture group and different culture groups. In the blue box, although both images belong to the Japanese cultural group, traditional Japanese clothing (i.e., kimono) is not described in the below image. In the red box are images from different cultural groups, but it is difficult to distinguish the group through the generated captions.}
    \label{fig:introduction}
\end{figure}

To date, many Vision-Language Pre-trained models (VLPs) demonstrate outstanding performance across various vision-language tasks~\cite{wang2022git,yu2022coca,li2023blip}. Despite the remarkable performance in image captioning through VLPs, the captions generated neglect to consider the cultural elements of different groups, such as the traditional clothing of the Asian people and cultural food for special occasions, etc. Most of these issues stem from their large-scale image-text pair training datasets, which are not necessarily collected with consideration for cultural diversity. Moreover, annotators, in general, may not pay attention to the cultural elements' details~\cite{wang2020compare}. Researchers have primarily used evaluation metrics~\cite{papineni2002bleu,vedantam2015cider} to estimate how closely generated captions resembling the annotations in the dataset, such as COCO~\cite{lin2014microsoft} and Flickr30K~\cite{plummer2015flickr30k}. However, these evaluation metrics pose a challenge in generating captions that describe cultural elements of cultural groups. as shown in Figure \ref{fig:introduction}.

In this paper, we introduce a framework that describes the cultural elements of given culture groups from images called \textbf{Cultural Image Captioning (CIC)}. CIC framework focuses on generating captions that include cultural visual elements perceived as cultural elements. Our goal is to generate culturally aware image captions about various cultural groups.

Our research question is, how can we extract cultural visual elements from existing image captioning models to generate culturally aware captions? One approach involves representatives from each cultural group collecting images and writing annotations to create benchmark datasets for cultural image captioning. However, recruiting representatives from each group demands a significant amount of human resources, and variations in the cultural understanding among representatives lead to diverse human bias. So, collecting benchmark datasets from various culture groups to train the model can be unrealistic.

Similar challenges are observed in the Visual Question Answering (VQA) regarding human annotations~\cite{banerjee2020weaqa,changpinyo2022all}. To address these aforementioned issues, zero-shot VQA approaches relying on Large Language Models (LLMs) have been proposed, given the demonstrated excellence of LLMs in zero-shot QA task~\cite{yang2022empirical,guo2022images}. The fundamental idea of these approaches is to integrate the visual modality into LLMs that typically handle only the language modality. \cite{guo2022images} generates captions for candidate answer regions and, based on these, creates question-answer prompts to serve visual information for zero-shot VQA to the LLMs.

Inspired by this zero-shot VQA approach, our framework involves (1) generating cultural questions based on cultural categories, (2) extracting cultural visual elements from VQA using generated questions, and (3) generating culturally-aware captions using LLMs with the prompts. As images may not contain cultural elements for all cultural categories, our framework proposes a method to extract representing cultural categories in images to prevent hallucinations in LLM. We construct a prompt for LLM to generate culturally-aware captions based on VQA results.

To evaluate our results, we recruited participants with a high cultural understanding for each of the four cultural groups defined in the GD-VCR~\cite{yin2021broaden} dataset: West, South Asia, Africa, and East Asia. Participants rated captions based on the quality of cultural descriptions among those generated by the baseline models and our framework for 10 randomly sampled images. The 450 responses from 45 participants demonstrate that our framework can generate higher-quality cultural descriptions than the baselines.
Our contributions are as follows:

\begin{enumerate}
    \item Introducing cultural awareness image captioning that depicts cultural visual elements by reflecting on the cultural categories of the images.
    \item Generating cultural questions for conducting Visual Question Answering (VQA), obtaining cultural visual elements, and transferring it to a Language Model (LLM) to generate culturally aware captions.
    \item Human evaluations by representatives from various cultural regions, our framework was deemed to generate captions that best describe images in a cultural context.
\end{enumerate}

\section{Related Work}
\subsection{Cultural Bias}
Previously, various methods have been proposed to address the cultural bias in AI applications~\cite{wang2022revise,mandal2021dataset,yin2022geomlama}. The Dollar Street~\cite{rojas2022dollar} dataset consists of indoor images collected from Western and non-Western countries based on socioeconomic features such as household items and monthly income. \cite{yin2023givl} utilized it as a benchmark dataset for a geo-diverse classification task. GD-VCR~\cite{yin2021broaden} and MaRVL~\cite{liu2021visually} datasets are developed to mitigate cultural bias in reasoning tasks. WIT~\cite{srinivasan2021wit} aims to address cultural bias in text-image retrieval from geo-diverse Wikipedia images. So far, no Image Captioning benchmark dataset has been presented that considers regional diversity. 

Other work has introduced VLPs trained on geo-diverse knowledge~\cite{yin2023givl}. However, they evaluated performance in common image captioning without considering cultural features. To our knowledge, our framework is the first method to consider the cultural elements of geo-diverse images in generating Culturally-aware Image Captioning.

\subsection{Vision-Language Pre-trained Models (VLPs) for Image Captioning}
VLPs~\cite{wang2022git,yu2022coca,li2023blip} have been proposed to address various Vision-Language multi-modal tasks. VLPs obtain Vision-Language representations from raw image-text pair data and have demonstrated high performance when applied to diverse V-L downstream tasks~\cite{long2022vision}. Among these tasks, image captioning is evaluated using benchmark datasets such as COCO~\cite{lin2014microsoft}, Flickr30~\cite{plummer2015flickr30k}, and Conceptual Captions~\cite{sharma2018conceptual}. However, these datasets are mostly based on Western-centric images, and their annotations consist of general words that do not consider cultural factors. Furthermore, the use of evaluation metrics like BLEU~\cite{papineni2002bleu}, CIDEr~\cite{vedantam2015cider}, and SPICE~\cite{anderson2016spice}, which assess how similar generated captions are to annotations. So, generating culturally-aware image captions with existing image captioning using VLPs is difficult.

\subsection{LLM for Zero-shot VQA Tasks}
The importance of the zero-shot VQA task has been emphasized to address the difficulty in generating VQA data~\cite{banerjee2020weaqa,changpinyo2022all,yuan2021language}. Recently, as Large Language Models (LLMs) have demonstrated high performance with zero data, zero-shot VQA tasks have relied on leverage LLMs~\cite{yang2022empirical,guo2022images}. In aforementioned studies, methods have been suggested to incorporate visual information into LLMs that originally handled natural language modalities only. For instance,~\cite{guo2022images}generates multiple Question-Answer samples from captions generated for areas related to the question and uses them as prompts. Inspired by the approach of~\cite{guo2022images}, we generate cultural questions and create Culturally-aware Image Captions using cultural visual information obtained from VLPs through questions.

\begin{figure*}
    \centering
    \includegraphics[width=0.85\textwidth]{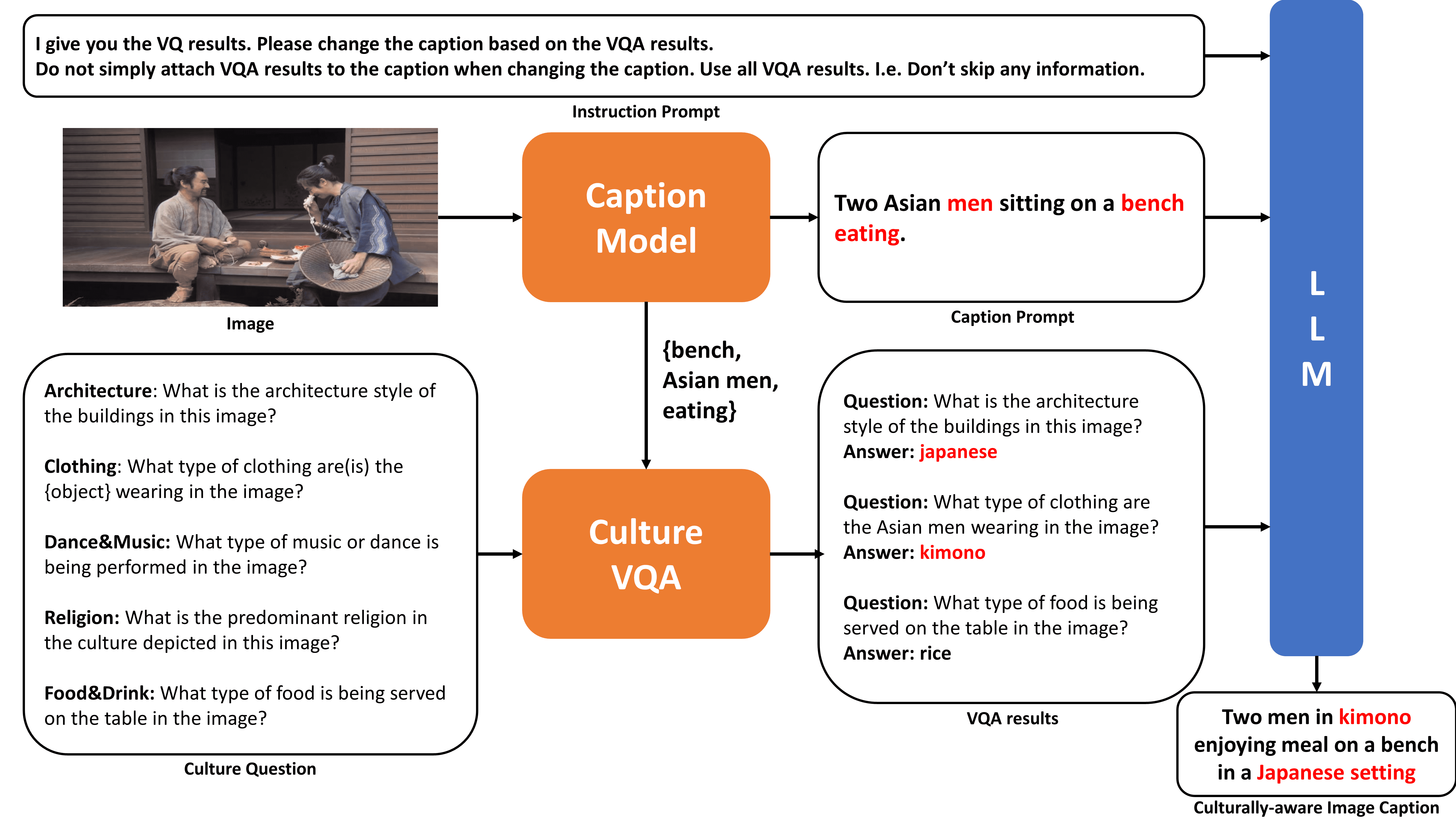}
    \caption{\textbf{CIC Overview.} First, culture questions are generated as described in Section ~\ref{sec:generate}. Then, cultural visual elements represented in the image are extracted through VQA as described in Section ~\ref{sec:extracting}. Finally, LLM generates culturally-aware captions as described in Section ~\ref{sec:promt}}
    \label{fig:framework}
\end{figure*}

\section{Method}
Our overall framework is depicted in Figure \ref{fig:framework}. In this section, We present the generation of cultural questions (Section ~\ref{sec:generate}), VQA for extracting cultural visual information (Section ~\ref{sec:extracting}), and Prompt design for cultural captions generation (Section ~\ref{sec:promt}).

\subsection{Generate Cultural Question}\label{sec:generate}
Based on the definitions of culture in~\cite{liu2021visually} and~\cite{halpern1955dynamic}, five categories are used to extract cultural elements from images: architecture, clothing, dance and music, food and drink, and religion. Following the process in~\cite{zhu2023chatgpt}, we created questions set $Q=\{q_j\}_{j=1, \dots, U}$, where U is the total number of questions to get the cultural elements based on cultural categories. We generated five questions per image in GD-VCR~\cite{yin2021broaden}, and U is 1,645. The objective of using cultural questions is to extract cultural visual elements from the image. So, we first filtered out questions that were not associated with cultural categories or images (e.g., ``What is the overall mood or atmosphere conveyed by the portrait?"). $Q'$ is the questions set after filtering:

$$Q'=\{q^c_j\}^{c}_{j=1, \dots, u}, u\leq U$$ 

where $c \in$ \{Architecture, Clothing, Food \& Drink, Dance \& Music, Religion\} is cultural categories; u is the total number of questions after filtering.

Subsequently, $Q'$ contains similar cultural questions for each cultural category. Similar questions may produce equal VQA results, so clustering similar questions using SentenceTransformer~\cite{reimers2019sentence}. Table ~\ref{tab:1} shows the results of removing irrelevant questions and questions clustering. We cluster questions based on 90\% similarity and set each cluster to have at least 8 questions. Multiple groups can be created for each category, and each group did not contain questions related to different cultural elements (i.e., question groups for architecture did not include questions related to other categories). The definition of a group is below: 

$$G^{c'}_{i \dots N} = \{q^{c} | q^{c} \in Q', c=c'\}$$ 

where $c' \in $ \{Architecture, Clothing, Food \& Drink, Dance \& Music, Religion\} is cultural categories; N is number of the cluster.

Lastly, we calculated the precision based on cultural elements and cultural groups for each cultural category from the VQA results of BLIP2 using the questions of each cluster for the final questions. For example, when the question ``What is the architectural style of the building in this image?" is asked in the image of the African cultural group, the answer ``kenya style" is set to TP, and the answer is ``modern" is set to FP. We determined cultural elements based on Culture Commonsense Knowledge (CCSK)~\cite{nguyen2023extracting}. We present the questions with the highest precision values for each category in Figure ~\ref{fig:framework}. We put precision results for each cluster's questions in the appendix.

\begin{table}
    \centering
    \begin{tabular}{ccc}
        \hline
         \textbf{category} & \textbf{\# Question} & \textbf{\# Cluster} \\
        \hline
         \textbf{Architecture}& 123 & 7\\
         \textbf{Clothing}& 203 & 9\\
         \textbf{Food \& Drink}& 225 & 4\\
         \textbf{Dance \& Music}& 351 & 6\\
         \textbf{Religion}& 152 & 5\\
        \hline
         \textbf{total} & 1,054 & 31\\
        \hline
    \end{tabular}
    \caption{Results of the number of questions and the number of clusters by category}
    \label{tab:1}
\end{table}

\subsection{Extracting Culture Elements from Cultural Visual Question Answering (VQA)}\label{sec:extracting} 
We extracted cultural visual elements from images through BLIP2's Visual Question Answering (VQA) using cultural questions generated in the previous section. However, cultural visual elements for the five cultural categories rarely appears in actual images at once. Therefore, when the VQA results for all cultural questions previously created are delivered to the Large Language Model (LLM), a hallucination problem may arise where the LLM may generate captions describing cultural elements not present in the images. To address this, if words related to cultural categories are generated in the caption, we assume that cultural elements is present in the image. To extract words related to cultural categories, we construct instructions for ChatGPT\footnote{https://openai.com/blog/chatgpt}, which has demonstrated excellent performance in zero-shot reasoning: ``Please extract the words related to Architecture, People, Food \& Drink, Dance \& Music, and Religion from Caption. Caption: [caption]". We extract words related to `People' for clothing to inquire about people's clothes. Then, we perform VQA based on the word categories extracted from LLM to obtain cultural visual information. In Figure ~\ref{fig:framework}, within the caption ``Two Asian men sitting on a bench eating.", `bench,' `Asian men,' and `eating' are extracted as words related to Architecture, Clothing, and Food \& Drink, respectively. The VQA results are obtained with questions related to these cultural categories.

\subsection{Prompt Design}\label{sec:promt} 
We construct a prompt for LLM to generate captions by concatenating the instruction, caption prompt, and VQA results. The instruction text is ``I will give you the VQA results. Please change the caption based on the VQA results. Do not simply attach the VQA results to the caption when you change the caption. Use all the VQA results. I.e., Don’t skip any information." The Instruction prompt directs LLM to generate the culturally-aware image caption based on the caption prompt and VQA results. For LLM to generate captions with one sentence, the phrase ``Do not simply attach the VQA results" was added to the instruction. Additionally, to prevent cases where LLM arbitrarily does not reflect the results of VQA, the phrase ``Use all the VQA results” was added. The caption prompt is ``Caption: [caption]" and provides LLM with information about the interactions between objects in the image. The VQA prompt is constructed by repeating ``Question: [question] Answer: [answer]" for the all of cultural categories represented in the image.

\section{Experiments}
\noindent \textbf{Dataset.} We validated our framework using GD-VCR~\cite{yin2021broaden}, a multiple-choice QA testing set designed to evaluate the ability of multi-modal models to understand geo-diverse commonsense knowledge. There is no benchmark dataset for Geo-diverse Image Captioning, so we selected GD-VCR to evaluate our framework. The dataset includes 329 images representing four cultural groups: West, East Asia, South Asia, and Africa.

\noindent \textbf{Baseline models.} We chose the following models as baseline models: (1) GIT~\cite{wang2022git}, a VLP for image-to-text generation tasks; (2) CoCa~\cite{yu2022coca}, demonstrating excellent performance in zero-shot image captioning with a multi-modal text decoder; and (3) BLIP2~\cite{li2023blip}, a VLP that performs well in various vision-language tasks.

\noindent \textbf{Implementation details.} To obtain caption prompts and captions for culture VQA, we use BLIP2 for caption generation. Through Table ~\ref{tab:4}, we selected BLIP2, a baseline model demonstrating good CLIP scores across various cultural groups. We used ChatGPT as the LLM for generating culturally-aware image captions. The temperature is set to 0.6, and the maximum length for caption generation is 100.

\noindent \textbf{User survey.} Our goal of improving cultural awareness of generated captions is subjective evaluation determined by people from the given cultural groups. To evaluate the performance of our framework based on these criteria, we recruited participants aged 18 years or older from each cultural group. Each participant responded to 10 survey pages. Each survey page has one image and two survey items. The first survey item identifies the match between the cultural categories selected by participants and the culture components by the VQA described in Section ~\ref{sec:extracting}. The second survey item lets the participants choose one among four given captions that describe the cultural components of a given image. The first survey item allows multiple selections, and the second is a single selection item. We quantitatively estimate the subjective perceived performance of each survey item. The details about our survey can be found in the appendix.  

\noindent \textbf{Automatic Metrics.} In addition to human evaluation, we use the CLIPScore~\cite{hessel2021clipscore}, a metric for evaluating captions in situations where there are no reference captions for the given captions. Additionally, we introduce a new evaluation metric called \textbf{Culture Noise Rate (CNR)} to measure the rate of cultural words in the generated captions, as described below.

\begin{figure*}
    \centering
    \includegraphics[width=\textwidth]{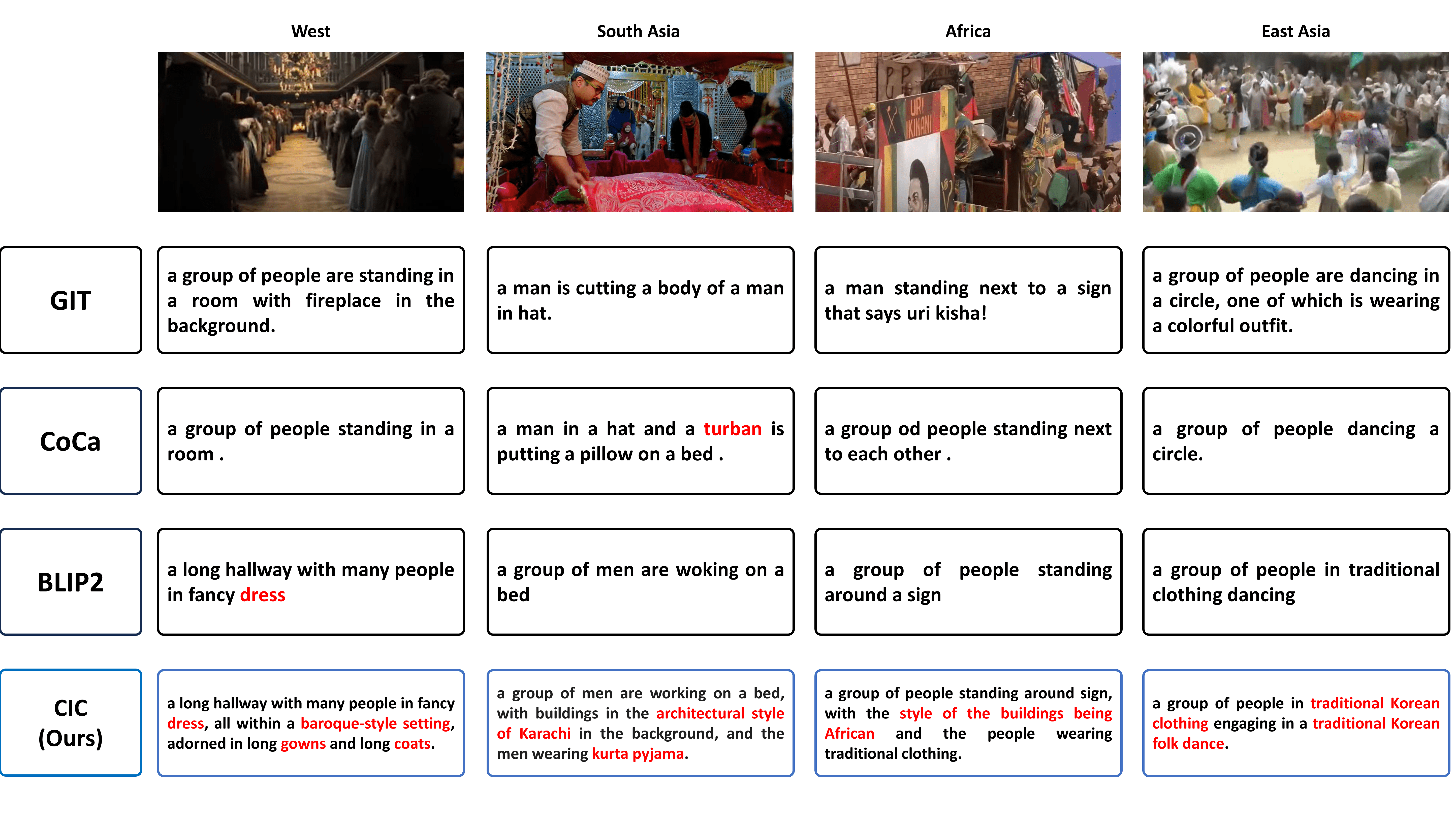}
    \caption{Caption generated for the image depicted in the given 4 different cultural groups by the baseline models and our framework (CIC) in the paper. The red words in each caption represent visual elements of each culture. Compared to existing baseline models, our framework describes images more culturally.}
    \label{fig:Qualitative}
\end{figure*}

\section{Results}
\subsection{Qualitative Comparison}  
Figure ~\ref{fig:Qualitative} shows the difference in the captions generated by four different models, including ours. CIC guides the generation of captions to describe cultural visual elements not captured in captions generated by existing VLPs. For example, while baseline models describe spaces without cultural visual elements, such as `hallway' or `bed,' CIC can describe the architectural style of the relevant cultural group. Additionally, CIC also describes the country to which the cultural group belongs.

\subsection{User Survey Results}

Improving the cultural understanding of the generated captions is primarily assessed subjectively by participants from the respective cultural groups. 45 participants from 4 different cultural groups in the GD-VCR~\cite{yin2021broaden} dataset responded to the survey. Each participant conducted a survey for 10 images and collected 450 responses.

\begin{table}
    \centering\resizebox{\linewidth}{!}{
    \begin{tabular}{ccccc}
    \hline
         \textbf{Architecture}&\textbf{Clothing}  &\textbf{Food \& Drink}  &\textbf{Dance \& Music}  &\textbf{Religion} \\
    \hline
         0.64& \textbf{0.96} & \textbf{0.78} & \textbf{0.84} & 0.64\\
    \hline
    \end{tabular}}
    \caption{Results of the rate of match between the cultural categories selected by participants and the culture components by the VQA described in Section ~\ref{sec:extracting}.}
    \label{tab:2}
\end{table}
\noindent\textbf{Culture Categories Match.} The first survey item evaluates how well the method of extracting cultural elements from the images described in ~\ref{sec:extracting} aligns with what people from the respective cultural regions perceive. Participants selected multiple cultural categories they believed were present in the images. Table ~\ref{tab:2} shows the matching rate between participants' responses and the cultural categories extracted by our framework. The existing image captioning models seem to focus more on the interactions between the objects in the image. Therefore, among cultural categories, clothing, food \& drink, and dance \& music, which are related to depicting interactions between objects, show a high matching rate. However, existing image caption models do not describe the surrounding background in much detail, so the user survey shows a relatively low matching rate for cultural visual elements such as architecture and religion.

\begin{table}
    \centering
    \begin{tabular}{cccccc}
    \hline
         Model & West & South Asia & Africa & East Asia\\
    \hline     
         GIT &  0.08& 0.1 & 0.17 &  0.16 \\
         CoCa&   0.1 & 0.21 & 0.17 & 0.15\\
         BLIP2&   0.2 & 0.14 & 0.21 & 0.26\\
    \hline
         \textbf{CIC (ours)}& \textbf{0.62} & \textbf{0.55} & \textbf{0.44} &  \textbf{0.43} \\
    \hline
    \end{tabular}
    \caption{Results of user survey based on the quality of cultural descriptions.}
    \label{tab:3}
\end{table}

\noindent\textbf{Culturally-aware Image Caption.} The second survey item evaluates how well our framework's generated captions depict cultural aspects compared to captions generated by existing image caption models. Participants rated captions based on the quality of cultural descriptions. Table ~\ref{tab:3} shows that our framework received significantly higher scores across all four cultural groups—West, South Asia, Africa, and East Asia—compared to other baseline models, ranging from approximately 17\% to 42\%. Based on user survey results, we can conclude that our framework excels in generating cultural descriptions from cultural images. 

Additionally, the results show a notable difference in the scores between the West, South Asia, Africa, and East Asia. We attribute this to the bias in the data used by existing VLPs. According to~\cite{ipeirotis2010demographics}, most annotators of the data used by the VLPs are from the United States and India. From this, we infer that VLPs were trained with a bias towards Western and South Asian cultures, leading to better cultural VQA results in these regions.

\begin{figure*}
    \centering
    \includegraphics[width=\textwidth]{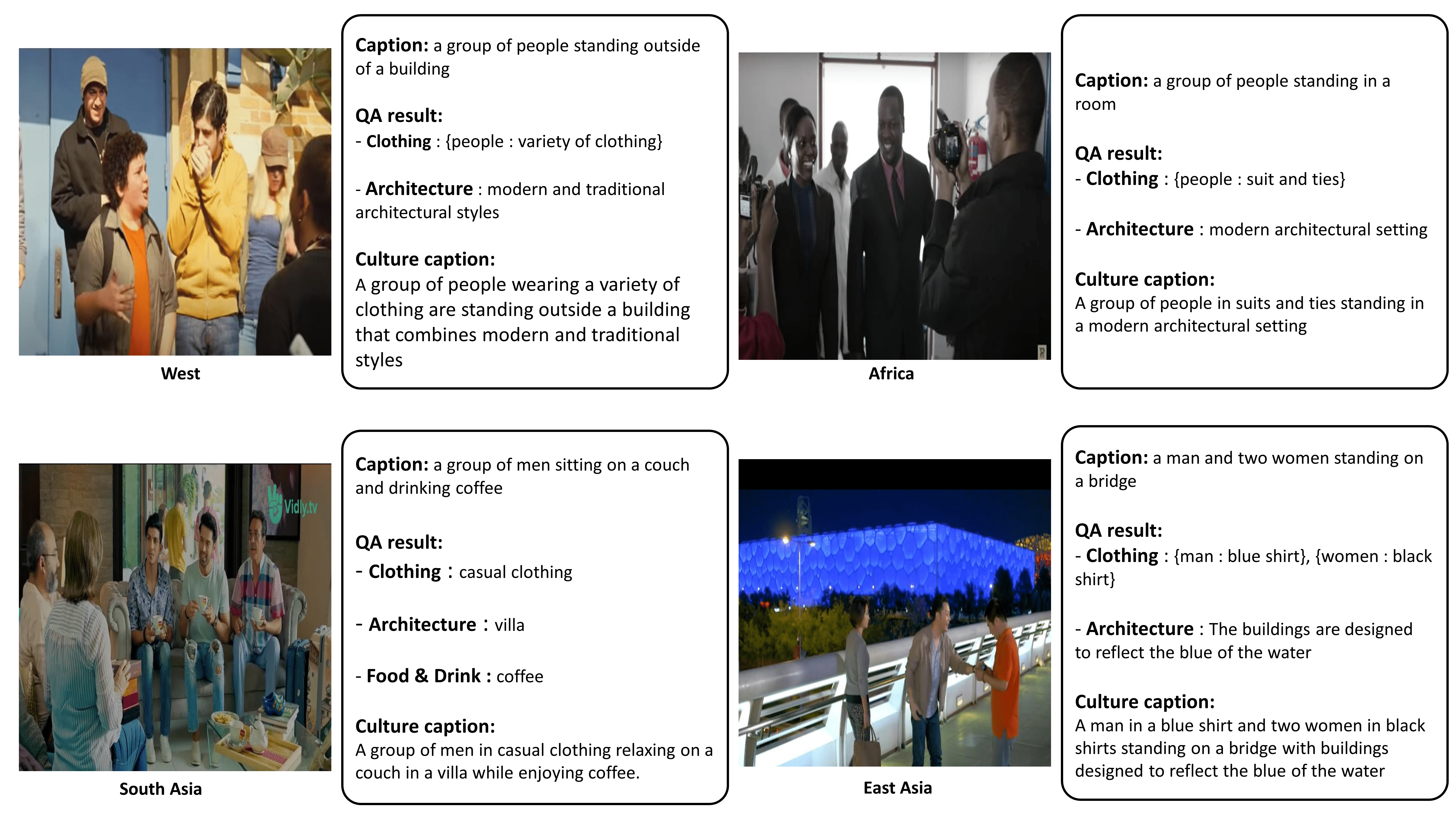}
    \caption{Caption generated for the modern cultural image depicted in given 4 different cultural groups by our framework. The words that specify cultural groups are not created in the generated culturally-aware captions, making it difficult to distinguish between cultures through only captions.}
    \label{fig:modern}
\end{figure*}

\begin{table}[ht!]
    \centering\resizebox{\linewidth}{!}{
    \begin{tabular}{cccccc}
    \hline
         Model & West & South Asia & Africa & East Asia & total\\
    \hline     
         GIT &  0.6939& 0.6571 & \textbf{0.6972}  & 0.6821 & 0.6824 \\
         CoCa&   0.6580 & 0.6402 & 0.6338  & 0.6749 &0.6594\\
         BLIP2&   0.6968 & 0.6659 & 0.6541  & 0.6920 &0.6821\\
    \hline
         \textbf{CIC (ours)}& \textbf{0.7109} & \textbf{0.6911} & 0.6808 &  \textbf{0.7168} & \textbf{0.7036}\\
    \hline
    \end{tabular}}
    \caption{Results of CLIPscore of the caption generated by baseline models and framework.}
    \label{tab:4}
\end{table}

\begin{table}[ht!]
    \centering\resizebox{\linewidth}{!}{
    \begin{tabular}{cccccc}
    \hline
         Model & West & South Asia & Africa & East Asia & total\\
    \hline     
         GIT &  1.688& 0.963 & 0.697  & 1.913 & 1.515\\
         CoCa&   1.994 & 1.149 & 0.25  & 1.962 &1.546\\
         BLIP2&   2.745 & 0.728 & 0  & 1.565 &1.362\\
    \hline
         \textbf{CIC (ours)}& \textbf{8.484} & \textbf{15.042} & \textbf{16.282} &  \textbf{19.008} & \textbf{17.142}\\
    \hline
    \end{tabular}}
    \caption{Results of Culture Noise Rate (CNR), which calculates the ratio of cultural visual elements in a sentence. In cultural groups excluding the West, ~15\% more cultural elements are depicted.}
    \label{tab:5}
\end{table}

\subsection{Automatic Metric Result}
Our use of GD-VCR, which was not originally designed for the image captioning task, lacks reference captions for evaluating generated captions. Therefore, we employ CLIPScore~\cite{hessel2021clipscore} as a metric to evaluate captions without reference captions. As shown in Table ~\ref{tab:4}, our framework shows high CLIPScore for three of four cultural groups, excluding Africa. The reason for the low CLIPScore for Africa is that our framework utilized captions generated by BLIP2 as the caption prompt. However, in the case of Africa, GIT outperformed BLIP2 in CLIPScore. Nevertheless, CLIPScore is an indicator that evaluates images and captions based on CLIP ~\cite{radford2021learning}, and having a high CLIPScore does not necessarily imply culturally descriptive captions.

Therefore, we introduce the\textbf{ Culture Noise Rate (CNR)} as a new evaluation metric for culturally-aware image captions. CNR is defined as the ratio of cultural words among all words generated by a caption. Cultural words are determined based on cultural commonsense Knowledge (CCSK)~\cite{nguyen2023extracting}. As shown in Table ~\ref{tab:5}, our framework generates approximately 17\% more cultural words than other baseline models. However, the West group exhibits a lower ratio of cultural words than other cultural groups. This means it is difficult for VQA to answer cultural elements for the Western images. 

\begin{table}
    \centering\resizebox{\linewidth}{!}{
    \begin{tabular}{cccccc}
    \hline
         Model & West & South Asia & Africa & East Asia & total\\
    \hline     
         \textbf{CIC (Ours)} &  \textbf{0.7109}& \textbf{0.6911} & \textbf{0.6808}  & \textbf{0.7168} & \textbf{0.7036}\\
         w/o caption prompt&   0.6162 & 0.6230 & 0.6080  & 0.6369 &0.6219\\
    \hline
    \end{tabular}}
    \caption{Results of CLIPScore with and without caption prompt.}
    \label{tab:6}
\end{table}

\subsection{Modern Culture Case Analysis}

In Figure ~\ref{fig:modern}, the results of culturally-aware image captioning for modern images in the four given cultural groups are presented. When looking at the four images, it is not difficult to distinguish between cultural groups. However, it is challenging to identify cultural elements from the generated captions, and distinguishing cultural groups based solely on the captions is not possible. This highlights the limitation of our defined five cultural elements, which can only differentiate unique cultures within a cultural group. For modern cultural images, considering additional cultural elements, such as race and modern city style, would be a direction for future work.

\subsection{Ablation Study on Prompt Design} \label{ablation}
When constructing the prompt for LLM, we provide captions generated by existing VLPs. This objective is when using only VQA results as prompts for generating captions, LLM has difficulty describing the interaction between objects. We conducted additional experiments by excluding the caption prompt and using only instructions and VQA results to compose the LLM's prompt. As shown in Table ~\ref{tab:6}, the design that provides captions shows significantly higher CLIPScore than the design that does not. To connect LLM's visual modality, we confirm that captions convey information about object interaction. Consequently, we confirm that captions provide LLM with information about the interaction between objects.

\subsection{Ablation Study on Vocabulary Extracted from Caption Prompts}
In our cultural questions regarding the clothing category, we replace \textit{\{object\}} with terms related to `people' such as `men', `women', `man', and `woman' extracted from the caption prompt. The purpose is to obtain detailed information about what each person is wearing in the image, especially in cases with multiple individuals. We conducted additional experiments replacing \textit{\{object\}} with `people' instead of using words extracted from the caption prompt: ``What type of clothing are the people wearing in the image?". As shown in Table ~\ref{tab:7}, using words extracted from the caption prompt results in a higher CLIPScore than using only 'people'. Based on these results, We infer that using words extracted from the caption prompt enables more detailed descriptions of individuals in Cultural Visual Question Answering and leads to a higher CLIPScore than not using terms related to `people'. Consequently, we confirm that using culture-related vocabulary extracted from the caption prompt enables the obtaining more detailed information.

\section{Limitation}
Our framework generates culturally-aware captions using culture-related visual information extracted from images, and we evaluate it using CLIPScore since there are no other appropriate measures for non-reference image caption data. However, CLIP model itself has biases ~\cite{birhane2021multimodal}, making it less suitable for judging cultural factors, and currently, our approach relies on human evaluation. A qualitative metric that can assess cultural elements from image-text pairs is needed for our future research.

Furthermore, as discussed in the results, the model falls short of capturing cultural elements as comprehensively as humans. This suggests the need for model training focusing on more diverse cultural elements depicted in images, which is proposed as a future work.

\section{Conclusion}
This paper proposes the CIC framework, aiming to generate culturally-aware captions for a given cultural group that are recognizable to individuals within that group. The goal is to create captions containing extracted cultural visual elements from images, contributing to cultural diversity. Our proposed framework has the potential for application not only in cultural domains but also in other domains, such as artworks and fashion, as an additional method for visual understanding. Extensive user surveys and metric evaluations demonstrate that our framework extracts cultural visual elements similar to what individuals from each cultural group perceive as cultural elements, resulting in more culturally descriptive image captions in subjective metrics.

\begin{table}
    \centering\resizebox{\linewidth}{!}{
    \begin{tabular}{cccccc}
    \hline
         Model & West & South Asia & Africa & East Asia & total\\
    \hline     
         \textbf{CIC (Ours)} &  \textbf{0.7109}& \textbf{0.6911} & \textbf{0.6808}  & \textbf{0.7168} & \textbf{0.7036}\\
         w/o vocab&   0.6752 & 0.6729 & 0.6672  & 0.7019 &0.6810\\
    \hline
    \end{tabular}}
    \caption{Results of CLIPScore with and without words extracted from caption prompt.}
    \label{tab:7}
\end{table}

\section*{Ethical Statement}
CIC framework was constructed using BLIP2 and ChatGPT, which trained on unfiltered data potentially laden with stereotypes or inherent bias. To ensure the evaluation of culturally-aware captions, we recruited participants from the respective cultural groups for user surveys. However, we did not evaluate the ethical aspects, such as cultural correctness and cultural/social biases in user surveys. Consequently, the proposed framework may replicate biases from the original model and should be considered stereotypes and biases in future work.

\section*{Acknowledgments}
This research was supported by the MSIT(Ministry of Science and ICT), Korea, under the ITRC(Information Technology Research Center) support program(IITP-2024-2020-0-01789), and the Artificial Intelligence Convergence Innovation Human Resources Development (IITP-2024-RS-2023-00254592) supervised by the IITP(Institute for Information \& Communications Technology Planning \& Evaluation).

\bibliographystyle{named}
\bibliography{ijcai24}

\end{document}